\documentclass{article}
\usepackage{spconf,amsmath,graphicx}

\usepackage{amsthm,amssymb}
\usepackage[ruled,vlined]{algorithm2e}

\title{Cartoondiff: Training-free Cartoon Image Generation with Diffusion Transformer Models}
\name{
  Feihong He$^{1}$, Gang Li$^{2,3}$, Lingyu Si$^{2}$,Leilei Yan$^{1}$, Shimeng Hou$^{4}$, Hongwei Dong$^{2}$, Fanzhang Li$^{\dagger,1}$\thanks{$^{\dagger}$Corresponding Author. \null\quad\  This work is supported in part by the National Key R\&D Program of China under Grant (2018YFA0701700, 2018YFA0701701), by the National Natural Science Foundation of China under Grant (61672364, 62176172, 62301539), and by the China Postdoctoral Science Foundation under Grant 2023M733615. },
}\address{
School of Computer Science and Technology, Soochow University$^1$,\\
  Institute of Software, Chinese Academy of Sciences$^2$,\\
  University of Chinese Academy of Sciences$^3$, \\
  Northwestern Polytechnical University$^4$}
\begin{document}
\maketitle
%
%
\begin{abstract}
Image cartoonization has attracted significant interest in the field of image generation. 
However, most of the existing image cartoonization techniques require re-training models using images of cartoon style. In this paper, we present CartoonDiff, a novel training-free sampling approach which generates image cartoonization using diffusion transformer models. 
Specifically, we decompose the reverse process of diffusion models into the semantic generation phase and the detail generation phase. Furthermore, we implement the image cartoonization process by normalizing high-frequency signal of the noisy image in specific denoising steps. CartoonDiff doesn't require any additional reference images, complex model designs, or the tedious adjustment of multiple parameters. Extensive experimental results show the powerful ability of our CartoonDiff. The project page is available at: https://cartoondiff.github.io/


\end{abstract}

\begin{figure*}[!t]
\centering
\includegraphics[scale=0.8]{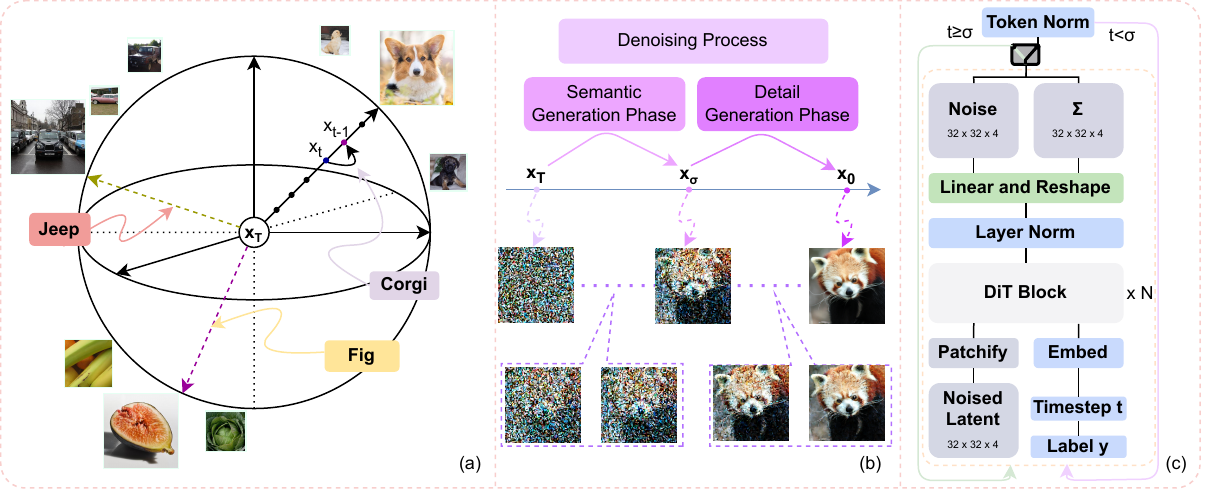}
\vspace{-2mm}
\caption{Methodology overview and model structure diagram. The diagram labeled (a) expresses the classifier-free extrapolation the models to specific classes under conditional guidance. The (b) illustrates our investigation of the denoising process for the diffusion model, categorizing it into the semantic generation phase and the detail generation phase based on the generated images for relative frequency information. In (c), we present the improvements made to the model based on the DiT\cite{DiT} structure. }
\vspace{-2mm}
\label{head}
\end{figure*}
\vspace{-2mm}
\section{Introduction}
\vspace{-1mm}
\label{sec:intro}

Cartoon-style art has experienced tremendous popularity and has been employed in various fields, e.g., animations, comics and games. Image cartoonization has attracted the attention of many researchers in the field of image generation. 
The current mainstream image cartoonization methods mainly are GAN-based~\cite{gan} methods.  In the previous stage, generative models were conventionally trained using paired datasets of real images and cartoon images~\cite{imageanalogies,pix2pix}. 
However, obtaining such paired data in real-world scenarios proves to be quite challenging. 
To address the challenge, CycleGAN~\cite{cycleGAN} and CartoonGAN~\cite{CartoonGAN} learn to translate between domains (real and cartoon images) without paired input-output examples. 
Furthermore, research efforts expand beyond the singular goal of generating cartoonized images, with a significant body of works making remarkable progress in various dimensions, e.g., interpretability~\cite{whitebox,closedgan,kejieshi}, the generation of diverse cartoon styles~\cite{MScartoongan} and guided cartoon generations~\cite{AniGAN,guidedGAN}. 

As one currently mainstream type of generative models, diffusion models~\cite{ddpm,ddim,LDM,li20223ddesigner} have gained widespread attention because of the stable convergence and diverse generation in visual content synthesis compared to GANs~\cite{gan}. Recently, Back-D~\cite{nulltext} represents the training-free cartoonization image generation using diffusion models, which employs rollback to perturb specific steps in the reverse process of classifier-free diffusion models. 

\begin{figure}[!t]
\centering
\includegraphics[scale=0.26]{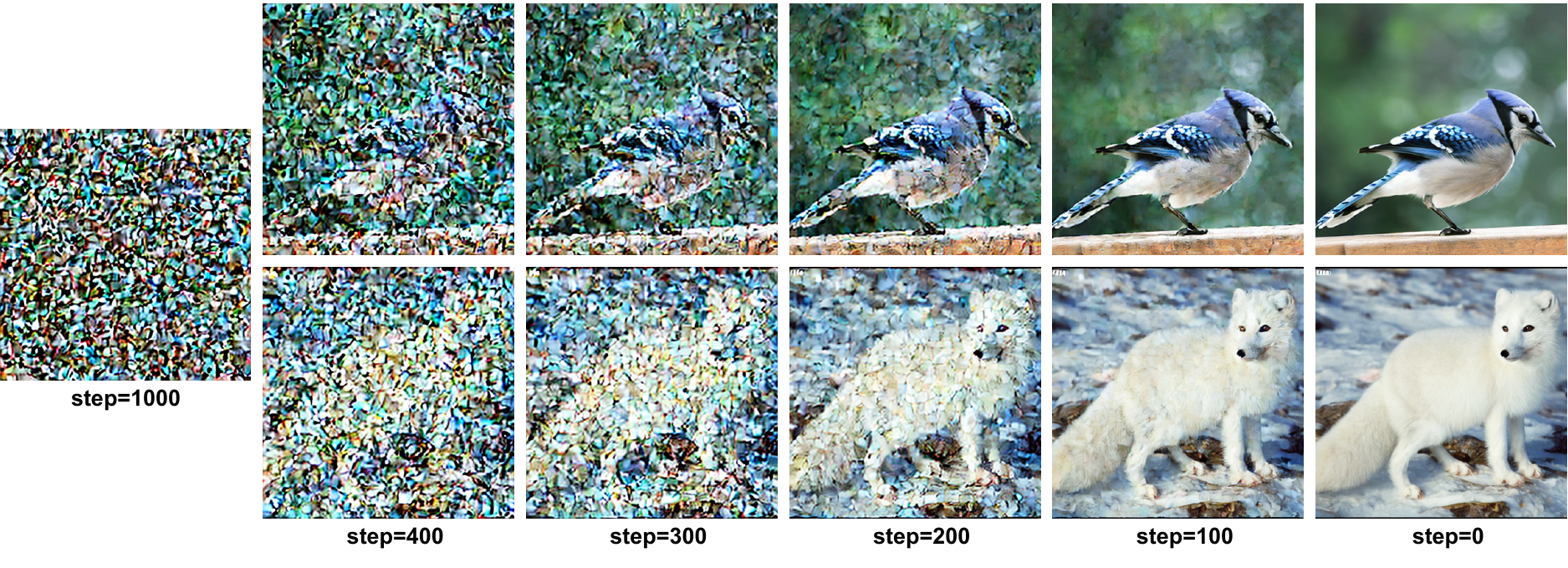}
\vspace{-3mm}
\caption{The intermediate results during the sampling process of DiT. }
\vspace{-3mm}
\label{midsteps}
\end{figure}

In this paper, we introduce a novel image cartoonization approach based on diffusion model called CartoonDiff. 
Compared to Back-D~\cite{nulltext}, our approach is simpler and more efficient. 
We first analyse classifier-free sampling and the denoising process in diffusion models, shown in Figure~\ref{head} (a) and Figure~\ref{head} (b), respectively. 
Figure~\ref{head} (a) clearly illustrates how classifier-free guided diffusion models align with different class directions under related class guidance. 
This ensures that the model generates images by class conditions during the denoising process. 
By exploring and analyzing this denoising process, we can broadly divide it into two stages: the semantic generation phase and the detail generation phase. 
As shown in Figure~\ref{head} (b), diffusion models build the overall semantic information of the image during the semantic generation phase and capture fine-grained texture details during the detail generation phase. 
To achieve the cartoonization of generated images, we employ a straightforward approach by inserting token normalization layers at specific steps during the denoising process to perturb the predicted noise, thereby restraining the generation of image textures and enhancing the image contours. 
Compared to our closest counterpart, Back-D~\cite{nulltext}, our method differs in that it eliminates the need for a complex rollback process, requires no additional image information, and entails minimal parameter adjustments. 
It is worth emphasizing that we are the first groundbreaking achievement in achieving training-free image cartoonization using the DiT~\cite{DiT} model. 
Despite its simplicity, it still exhibits superior performance. 
In summary, our work makes the following contributions: 

\begin{itemize}

\item Our work involves a comprehensive exploration and analysis of the denoising process in diffusion models. We discover that the diffusion model generates image semantic information in the early stages and detailed information in the later stages during the reverse process. 

\item Based on the aforementioned observations, we introduce a new training-free cartoonization method named CartoonDiff. It achieves image cartoonization by adding disturbances at specific steps during the reverse process of the diffusion model. 

\item We have achieved superior generation results compared to the concurrent training-free method, exemplified by Back-D, which is applied to stable diffusion models.




\end{itemize}

\vspace{-4mm}

\begin{figure*}[!t]
\centering
\includegraphics[scale=0.5]{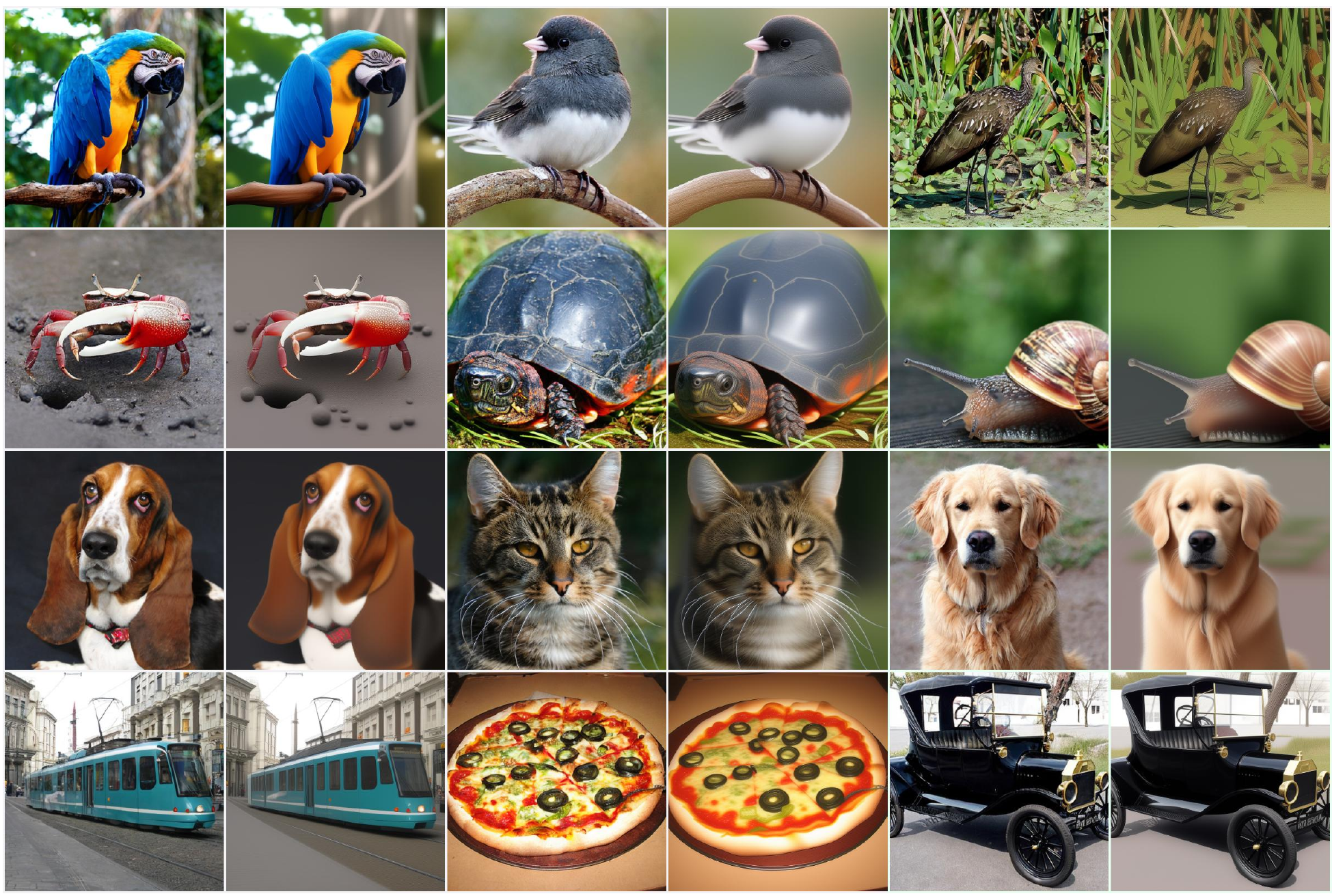}
\vspace{-3mm}
\caption{Based on DiT XL/2\cite{DiT}, with a hyperparameter $\sigma$ set to 250, we present the results of DiT's generation and the results generated using CartoonDiff. In each pair of images, the left side shows the original image generated by DiT, while the right side shows the image cartoonized using CartoonDiff. }
\vspace{-3mm}
\label{resultimgs1}
\end{figure*}

\vspace{-2mm}
\section{Method}
\vspace{-2mm}
\label{sec:format}
\subsection{Preliminaries}
\vspace{-1mm}
In the diffusion models, the forward process involves gradually adding noise to the image, resulting in a fully gaussian-noised image. This process can be represented as follows: $X_t = \sqrt{\Bar{\alpha}_t}X_0+\sqrt{1-\Bar{\alpha}_t}\epsilon$, where $\Bar{\alpha}_t$ is $\prod_{t=0}^s\alpha_t, \alpha_t\in\left(0,1\right)$ and $\epsilon$ stands for the result of random sampling from a gaussian distribution $N\left(0,I\right)$. 
In the reverse process of the diffusion model, noise prediction can be formulated using Bayes' theorem and the Markov property as follows: $P\left(X_{t-1}|X_{t},X_{0}\right) = \frac{P\left(X_{t}|X_{t-1},X_{0}\right)P\left(X_{t-1}|X_{0}\right)}{P\left(X_{t}|X_{0}\right)}$. 
The diffusion models optimize by measuring the KL divergence between the forward noise and the predicted noise distributions and constructs its variational lower bound$\left(VLB\right)$ to obtain the optimization objective as follows:
$Loss_t\left(\theta\right)=E_{t\sim\left[1:T\right],X_0,\epsilon_t}\left[\parallel \epsilon_t-\epsilon_{\theta}\left(\sqrt{\Bar{\alpha}_t}X_0+\sqrt{1-\Bar{ \alpha}_t} \epsilon_t,t \right) \parallel^2 \right]$.

Classifier-free guidance addresses the issues of additional training costs, classifier dependency, and adversarial attack effect with classifier guidance, making it widely adopted in numerous diffusion models. 
We utilize class information as conditional guidance, denoted as $c$. For noise prediction at step $t$, we can represent it as follows: $ \epsilon_\theta\left(X_t|c\right)=\epsilon_\theta\left(X_t|\emptyset \right)+\lambda\left(\epsilon_\theta\left(X_t|c\right)-\epsilon_\theta\left(X_t|\emptyset \right)\right) $. 
$\epsilon_\theta\left(X_t|\emptyset \right)$ represents the noise generated in the null-class condition $\emptyset$ within the classifier-free approach, while $\epsilon_\theta\left(X_t|c\right)$ represents the noise generated under class conditional guidance $c$. This formula facilitates the model's extrapolation for class-conditional image generation. 

\vspace{-2mm}
\subsection{CartoonDiff}
\vspace{-1mm}
To further analyse the diffusion model's sampling process, we visualize the intermediate results of different denoising steps. 
Figure~\ref{midsteps} showcases the results obtained through model denoising at different steps during the inference process, specifically at steps 1000, 400, 300, 200, 100, and 0, with a total of 1000 DDPM steps. 
It's evident that during the denoising process, the images first capture semantic information and then gradually acquire the finer details. 

\begin{algorithm}[!t]  
\label{CartoonDiffAlgorithm}
\caption{Training-free generation of the cartoon-style image with CartoonDiff}  
\SetAlgoLined
\LinesNumbered  
\KwIn{A pre-trained diffusion model $\epsilon_\theta\left(\cdot\right)$, conditional guidance $c$, guidance scale $\lambda$, disturbance time $\sigma$, temperature parameters $\Bar{\alpha}_t$.}
\KwOut{the cartoon-style image $X_0^*$}
\textbf{Initial: }$X_T\sim\mathbb{N}\left(0,I\right)$\;
\For{$t$ from $T$ to $0$}{
    $\epsilon_t = \epsilon_\theta\left(X_t|\emptyset \right)+\lambda\left(\epsilon_\theta\left(X_t|c\right)-\epsilon_\theta\left(X_t|\emptyset \right)\right)$\;
    \If{$t < \sigma$}{
        $\epsilon_t=Norm\left(\epsilon_t\right)$
    }
    $X_{t-1}= \sqrt{\Bar{\alpha}_{t-1}}\left(\frac{X_t-\sqrt{1-\Bar{\alpha}_t}\epsilon_t}{\sqrt{\Bar{\alpha}_t}}\right)+\sqrt{1-\Bar{\alpha}_{t-1}}\epsilon_t$
}
\Return{$X_0$ as $X_0^*$}

\end{algorithm}

We analyze the experimental results concerning the denoising process in Figure~\ref{midsteps} and make minimal modifications to the DiT architecture to generate cartoon-style images. 
 As illustrated in Figure~\ref{head}(c), we incorporate a token normalization block into the output of DiT, performing normalization on tokens at specific steps to inject cartoon-style statistics:
\vspace{-1mm}
\begin{equation}
\small
    Norm\left(v_{token}\right)=\frac{v_{token}}{\max \left(\|v_{token}\|_{1}, \epsilon\right)}
    \label{equation1}
\end{equation}
 We achieve image cartoonization by normalizing the predicted noise in the latent space to suppress the generation of fine texture details. 
In order to preserve low-frequency signals such as image contours and lines, we maintain the diffusion model's ability to capture important image information by setting the hyperparameter $\sigma$. 
The denoising process is divided two stages: 
\vspace{-1mm}
\begin{equation}
\small
    \left\{
\begin{aligned}
 & \epsilon_\theta\left(X_t|\emptyset \right)+\lambda\left(\epsilon_\theta\left(X_t|c\right)-\epsilon_\theta\left(X_t|\emptyset \right)\right) ,t\geq\sigma \\
 & Norm\left(\epsilon_\theta\left(X_t|\emptyset \right)+\lambda\left(\epsilon_\theta\left(X_t|c\right)-\epsilon_\theta\left(X_t|\emptyset \right)\right)\right) ,t<\sigma \\
\end{aligned}
\right.
\end{equation}

the hyperparameter $\sigma$ indicates that we introduce perturbations to the generated noise starting at step $\sigma$ during the inference process and continue until step 0.


In Algorithm~\ref{CartoonDiffAlgorithm}, we provide a brief algorithmic overview of CartoonDiff. We perform token normalizing on the generated noise during the denoising process when $t<\sigma$ to achieve image cartoonization. 

\begin{figure}[!t]
\centering
\includegraphics[scale=0.9]{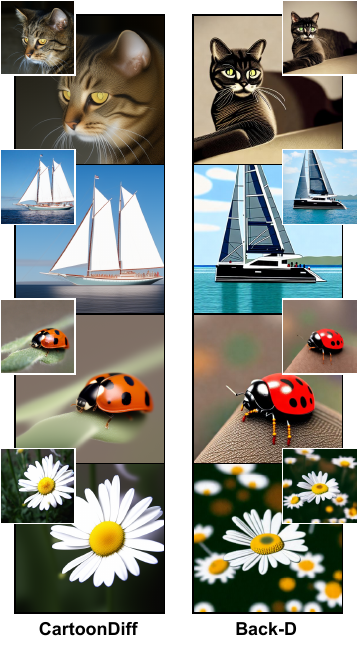}
\vspace{-5mm}
\caption{The comparative experiment between CartoonDiff and Back-D\cite{nulltext} includes images where each small image in the upper-left corner represents the original image output, while the larger image represents the output after cartoonization.}
\label{contrativeimg}
\end{figure}

\vspace{-2mm}
\section{EXPERIMENTS}
\vspace{-2mm}

In this section, we demonstrate the effectiveness of CartoonDiff through experimental results. 
We conducted experiments using the DiT-XL/2 model~\cite{DiT} pre-trained on the ImageNet $512\times512$~\cite{imagenetdataset} as the foundational generative network. 
DiT-XL/2 is a transformer architecture network comprising 28 layers of transformer blocks with the patch size of $2\times2$. 
During the inference phase, we simply append token normalization modules after its transformer blocks~\cite{transformer} to introduce perturbations at specific steps in the inference process, as illustrated in Figure~\ref{head}(c).

\subsection{Comparasion of generation Results}

 We adopt equidistant sampling with a sampling step setting of 100 (the total steps is 1000), while the hyperparameter $\sigma$ was configured as 250. In Figure~\ref{resultimgs1}, we present the generated results of cartoon images by our method applied to the DiT-XL/2 pre-trained model.
From the experimental results, we successfully achieve image cartoonization while preserving the essential details of the images. 
Additionally, we can notice that the cartoonized images exhibit diverse cartoon styles, because of subtle variations in their original art styles. For instance, the cartoonization results for ``limpkin" and ``basset" in the images exhibit a flat cartoon style, whereas the cartoonization results for ``parrot", ``crab", and ``golden retriever" display a more three-dimensional cartoon style. 

Additionally, we conducted a comparison between CartoonDiff and the existing training-free cartoonization method based on diffusion models. 
In the context of Back-D, we adhere to the recommended hyperparameter settings, with both hyperparameters b and s set to 300. Likewise, when configuring CartoonDiff, we maintain the aforementioned parameterization, specifying the hyperparameter $\sigma$ as 250. 

In Figure~\ref{contrativeimg}, we present comparative experimental results between Back-D and CartoonDiff. From the visualization, it is evident that our approach consistently achieves the desired cartoonization results. Conversely, Back-D tends to excessively enhance image contour information during the cartoonization process, resulting in a sense of dissonance in certain images, as exemplified by the ``cat" and ``daisy" images in the figure~\ref{contrativeimg}. 
It's worth noting that Back-D's core generative model is trained on the large-scale dataset LAION-5B~\cite{LAION}. 
And our model is exclusively trained on ImageNet, which does not include cartoon images. 

\begin{figure}[!t]
\centering
\includegraphics[scale=0.52]{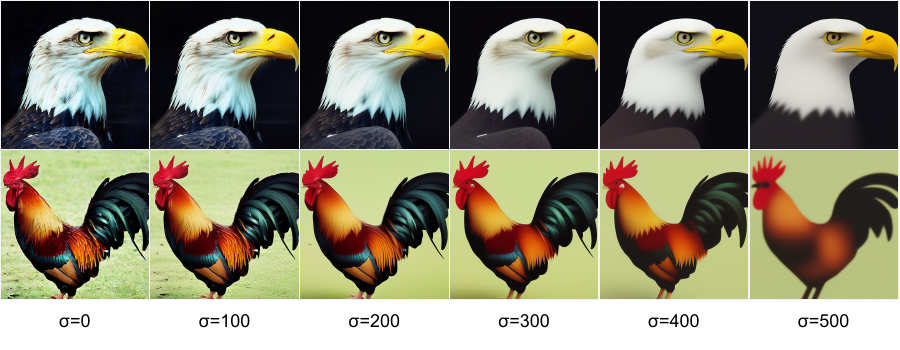}
\vspace{-3mm}
\caption{We adjusted the hyperparameter $\sigma$ to examine its impact on the significant image details and the degree of cartoonization. }
\vspace{-3mm}
\label{stepimg}
\end{figure}

\subsection{Ablation Study}

We perform  ablation experiments of hyperparameter $\sigma$ for our proposed method in Figure~\ref{stepimg}. We observe that as $\sigma$ increased, image details gradually smoothed and eventually disappeared. Simultaneously, within the $\sigma$ range of 0 to 400, low-frequency image information become both smoother and more pronounced. Based on our extensive experimentation, we find that $\sigma$ between 200 and 300 produce the best results.

\section{CONCLUSIONS}

In this paper, we first explore and analyze the classifier-free sampling and the denoising process in diffusion models. We discover the denoising process can be roughly divided into two phases, the semantic generation phase and the details generation phase. Based on that, we introduce the CartoonDiff method, which normalizes high-frequency details of the noisy image in specific denoising steps (the details learning phase). Experimental results show the effectiveness of our CartoonDiff. It is worth noting that, to the best of our knowledge, CartoonDiff is currently the simplest and most effective training-free method among cartoonization approaches based on diffusion models.


\vfill\pagebreak

\bibliographystyle{IEEEbib}
\bibliography{strings,refs}

\end{document}